# SDW-ASL: A Dynamic System to Generate Large Scale Dataset for Continuous American Sign Language

*Yehong Jiang*

The Nueva School

`yehjian@nuevaschool.org`

## Abstract

Despite tremendous progress in natural language processing using deep learning techniques in recent years, sign language production and comprehension has advanced very little. One critical barrier is the lack of largescale datasets available to the public due to the unbearable cost of labeled data generation. Efforts to provide public data for American Sign Language (ASL) comprehension have yielded two datasets, comprising more than thousand video clips. These datasets are large enough to enable a meaningful start to deep learning research on sign languages but are far too small to lead to any solution that can be practically deployed. So far, there is still no suitable dataset for ASL production.

We proposed a system that can generate large scale ASL datasets for continuous ASL. It is suitable for general ASL processing and is particularly useful for ASL production. The continuous ASL dataset contains English labeled human articulations in condensed body pose data formats. To better serve the research community, we are releasing the first version of our ASL dataset, which contains 30k sentences, 416k words, a vocabulary of 18k words, in a total of 104 hours. This is the largest continuous sign language dataset published to date in terms of video duration. We also describe a system that can evolve and expand the dataset to incorporate better data processing techniques and more contents when available. It is our hope that the release of this ASL dataset and the sustainable dataset generation system to the public will propel better deep-learning research in ASL natural language processing.

# 1. Introduction

According to a deaf community survey study conducted by R. Mitchell [1], 5% of the worldwide population suffers from hearing impairment. 80% of all deaf people cannot read or write. Here in the United States, about one million people are legally deaf, which is 4% of the population. Even among the educated deaf population, reading capability is far behind the performance of the hearing population. The mean reading grade level for a deaf high-school graduate is 5.9, while hearing people have a mean reading grade of 9.8. Additionally, there are 500,000 people who use American Sign Language (ASL) as their primary method of communication. English is usually a second language for most ASL users. According to the National Association of the Deaf, many deaf people feel that captions are inaccurate, difficult to follow, or "inaccessible" [2]. During the global COVID-19 pandemic, the federal and state governments, including the Department of Health [3], put up special effort to serve the deaf community by putting out ASL videos and hiring ASL signers for COVID-19 responses and other related announcements. However, many deaf people have still reported that they are unable to understand from coronavirus briefings what they need to do to stay safe and healthy [2].

Non-sign language users are enjoying better language related services than sign language users, such as machine command and dialogue, hands-free communication with devices, language translation, automatic caption, transcription and translation for streaming and conferencing video, etc. Some very early research works in sign language processing show that recent breakthroughs in machine learning, computer vision and natural language processing can also be applied to sign languages [9, 10, 11]. Advances in sign language research will eventually help to bring these services for non-sign language users to the Deaf community. In addition, computers may perform sign language interpretation when a human interpreter is not available.

It is known that deep neural network techniques require very large datasets. Lack of public-domain large sign language data has been recognized as the primary barrier against advancing sign language processing research. Efforts to provide public data for ASL understanding have yielded two datasets up to 80 hours of video clips. These datasets are large enough to enable meaningful early deep learning research on sign languages but are far too small to lead to any solution that can be practically deployed. So far, there is still no suitable dataset for ASL production.

In this paper, we introduce the SDW-ASL system that facilitates the generation of large scale ASL dataset and the release of the first generation SDW-ASL dataset to the public.

# 2. Background and Related Work

## 2.1. Methods to Record Sign Language

Sign language is a multimodal visual language. One way to record a sign language is through glossing. Gloss is a written approximation of another language. It is possible to gloss American Sign Language (ASL) using English with additional symbols. However, the gloss annotation requires human ASL linguists. The process is limited, time consuming and expensive. For example, How2Sign dataset includes gloss annotation, which took on average one hour per 90 seconds of video. Due to prohibitive cost, only part of the How2Sign dataset contains gloss annotation.

Here is an example of ASL gloss for an English sentence.

> English: *Are you busy two weeks from now?*
> ASL: *two-WEEK-future SATURDAY IX-you # BUSY?*

This gloss example illustrates that ASL, as an independent natural language, is different from spoken English. It is not a visual version of English. ASL has its own vocab and grammar rules. Besides the difficulty of gloss annotation, it should also be noted that gloss is still an imprecise approximation of the rich visual actuation of ASL.

Naturally, researchers record live video of human signers with English annotation. Raw video clips are clearly not a condensed form of any language. Most of the data in the video is not relevant to human action. Building a large dataset containing raw video not only takes huge storage space, but also won't be of any use to assist language production.

## 2.2. Sign Language Production and Comprehension

Language production is a form of action, and language comprehension is a form of action comprehension. Computer-based natural language processing includes domains like language production speech synthesis), language translation (English to German) and language understanding (machine dictation and conversation).

The production of multimodal visual sign language is much more difficult than speech synthesis. Speech synthesis is a mature field, often referred to as text to speech. We now take it for granted that machine generated voice from cloud services, like Amazon's Alexa, Apple's Siri, Google's Google, and Microsoft's Cortana, sound very natural. For speech synthesis, there is a simple 1:1 relation between a written word and its corresponding vocal sound. However, ASL production from English text is different from speech synthesis as it involves a language translation step followed by language synthesis step. Traditional research in ASL production does apply these two steps, from English text to ASL gloss to ASL actuation. Due to the above-mentioned challenges with ASL gloss annotation, previous research has had very little success.

A more practical research direction bypasses the intermediate ASL annotation stage and treats the English text to ASL actuation as a combined language translation process where an English word, phrase, or sentence is translated directly into the ASL multi-modal action sequence. Deep learning techniques such as Transformer Neural Networks are a very promising approach for this type of task. Success of any deep learning research relies on the confluence of three things: compute power, algorithm, and dataset. The first two are already in place. What's missing for ASL production research is a high-quality dataset. The creation of ImageNet back in 2009 procured the greatest come-back of neural-network-based Artificial Intelligence research. This is the primary motivator for the author to investigate ASL dataset generation.

## 2.3. Existing Datasets

So far, research publications of deep-learning-based sign language processing are primarily using the PHOENIX14 dataset [7] and its extension PHOENIX14T [8]. These datasets are in German Sign Language (DGS) and manually annotated in German. As shown in Table 1, PHOENIX14T has a moderate size of 11 hours and a very limited 3k vocabulary related only to weather forecasts.

MS-ASL [4] focuses on ASL vocabulary, not for continuous ASL sentences. As it doesn't contain any ASL grammar rules directly or indirectly, this dataset cannot be used for sign language production or comprehension. Studies have shown that simple word by word translation from English text to ASL actions result in very unpleasant experiences for deaf people. This can be understood from the above mentioned ASL gloss example. MS-ASL uses Internet sourced video clips, mainly based on ASL educational websites. Due to diverse video sources, the annotation process is still mostly manual. The process is tedious but less complicated as the goal is to extract only the single word or ASL vocab token. The token may be inferred based on the file name, or the web page tagging, or from the closed caption in the video clip.

How2Sign [2] is by far the largest ASL dataset available in public. It contains 80 hours of continuous ASL signing created in a controlled environment. One limitation of such an approach is that it cannot scale. It is an expensive content creation process with specific camera gear settings and hired ASL signers performing. To put it into perspective, How2Sign is based on a much larger natural language dataset called How2 [3]. Due to the cost and complexity, out of available 2000 hours of How2 contents, only 4% is reproduced for ASL in the How2Sign dataset. Additionally, 11 ASL signers participated in the recording. The creators of How2Sign did not attempt to assess its applicability to ASL production. There is no quantitative measure indicating the quality, dialect, and consistency of the signing.

Both MS-ASL and How2Sign were recently released to the public in 2021. There has not been any research community feedback yet.

| Dataset | Language | hours | videos | sentences | Words | vocab | signers | Annotation |
|---|---|---|---|---|---|---|---|---|
| PHOENIX14T | DGS | 11 | 8257 | 8257 | 114k | 3k | 9 | Manual |
| MS-ASL | ASL | 25 | 25513 | n/a | 25k | 1k | 222 | Semi-Auto |
| How2Sign | ASL | 79 | 35191 | 35191 | 598k | 16k | 9 | Manual |
| **This work** | **ASL** | **104** | **29626** | **29626** | **416k** | **18k** | **4** | **Auto** |

Table 1. Comparison of recent large public sign language datasets

## 3. Proposed Method

### 3.1. Dataset Requirements for ASL Language Production

Here are some key requirements for a great ASL dataset for deep learning based ASL production research.

- **Uniformity.** For language comprehension, it is important for the dataset to cover a variety of dialects, accents, genders, ages, etc. However, for language synthesis, uniformity is more important and is in fact a necessity.

- **High fidelity.** For ASL language comprehension, the dataset must contain live contents in a natural setting, with different lighting styles, backgrounds, signers, viewpoints, resolutions, noises, etc. For ASL production, the dataset needs to provide high quality, and to be easy to recognize and understand.

- **Controlled camera view.** For ASL language comprehension, a frontal view is preferred with the upper body fully exposed, and arms and hands (particularly the action arm and hand) fully visible all the time.

- **3D data capture.** ASL actuation are naturally three dimensional. Ideally, 3D data capture is desired. However, direct 3D data capture is prohibitively expensive for a large dataset. Recently, deep learning techniques have improved the quality of 3D action (gesture) produced from 2D video. These techniques can be used to generate 3D action data. As such technology continues to advance, it is important for the dataset generation tool to be adaptive to incorporate the new 3D action generation tools.

- **A large dataset with many words and rich vocabulary.** As the recent natural language processing (NLP) has shown, the larger the model, the larger the dataset, the higher fidelity the NLP neural network can achieve. For example, the GPT-3 model with 175 billion parameters was trained using the ever-growing Common Crawl dataset with close to 1 trillion words. The largest dataset for ASL is dwarfed in comparison — so far it is only around 10-thousand words. Dataset for ASL production needs to be large but doesn't have to match that for language comprehension. The dataset size should be at least in the 10's of thousand words and should have a path to grow into millions.

### 3.2. Proposed Dataset and Methodology

There are two objectives of this project. First, we want to create a publicly available ASL dataset serving as a standard for the ASL research community. It helps with research, paper publication and quantifiable comparison of research results. So far, most ASL dataset are designed for ASL understanding, neglecting a proper dataset for ASL production, which was the original motivation of the author. Secondly, to further propel AI research on ASL, we would like to design a system that allows the dataset to be evolved, expanded, and improved. In order to advance any ASL language model, the dataset has to be able to continue to grow.

#### 3.2.1. Crowdsourcing Content Selection

For ASL presentation purposes, signer divergence is not important. What is important is the fidelity of the data. Therefore, the signing must be consistent. Special dialects should not be included. If included, all signers should be using the same dialect. Secondly, the ASL vocab should be rich and scalable.

With the above-mentioned dataset requirement for ASL language production in mind, online news channels were selected as the content source. These news contents are typically 8 minutes long. Setting and video position of the news anchors are stable and consistent. All these videos contain a frontal view of right-handed anchors with news contents showing on the right side of the anchor. The ASL signing is mostly with the right hand with a clear view of the elbow above. All videos have stationary backgrounds. All anchors have a stable upper body position with minimal movements and no position change.

One important feature and a critical requirement for an ASL video clip is the availability of closed captions. Closed captions are time synchronized with the ASL video. It allows software automation to partition a video clip into smaller video segments - sentence by sentence, phrase by phrase for video clips with separate closed caption metadata files in a WebVTT (.vtt) format. WebVTT, standing for Web Video Text Track, is a popular subtitle and caption file format that contains the closed caption sentence data with corresponding start and end time. It is then used to split the video clip into ASL sentence segments with associated English sentences in text format. Some video clips are mastered with closed captions embedded in the video. For these clips, an OCR (optical character recognition) software is employed to detect the presence of closed captions and extract them. Time stamps need to be generated to segment the video clip. After several rounds of evaluation, Python-tesseract, which is a wrapper for Google's Tesseract-OCR Engine, was selected as the OCR software. Careful image pre-processing is conducted to ensure high quality OCR operation. Even though OCR results are mostly effective, there are still glitches once a while in dealing with scene transition, non-verbal closed caption, and the instability of closed capture bounding box.

Post processing of close capture data, including timestamps, are needed before applying it for video segmentation. For example, due to window size limitation for closed captions, a sentence could be divided into several fragments. Post processing was applied to combine these fragments into a full phrase or full sentence and combined the timestamps to form a sentence video segment.

### 3.2.2. Data Reduction

It might seem intuitive that a text-video pair (a time stamped English sentence text and ASL video segment) can form the basis for an ASL dataset. However, the video segment would have different backgrounds (that may change during the segment), different signers and different clothing. These background differences do not contribute to the ASL signing action. They are noises. Video segmentation and background removal could be applied to remove extraneous information while keeping the signer. However, there is no universally robust segmentation algorithm that we can rely on. As a visual language, ASL relies almost exclusively on the upper body actions with a focus on hands (detailed finger motion), arm, and facial expression. Therefore, ASL can be fully described by a high-fidelity upper body pose sequence. The upper body pose includes the positions for the head, body, arm and hands, face masks, and detailed hand and finger landmarks. Specifically, we use the first 24 landmarks of the BlazePose's 33 landmarks. 21 detailed hand finger landmarks are used which is very important for finger spelling. To record the detailed facial expression, we need to carefully track the movements of eyes, eyebrows, and mouth. In total, 468 three-dimensional face mesh landmarks are tracked.

Both OpenPose and MediaPipe were evaluated. We chose MediaPipe for its quality and speed. Based on MediaPipe, we developed a routine that combines post detection, face detection and hand detection. Using bounding box and outlier filtering, the robustness is greatly improved without sacrificing speed. As shown in Figure 1, the quality of finger tracking is good enough to recognize finger spelling.

By only releasing the post data sequence, the privacy of the signers is protected.

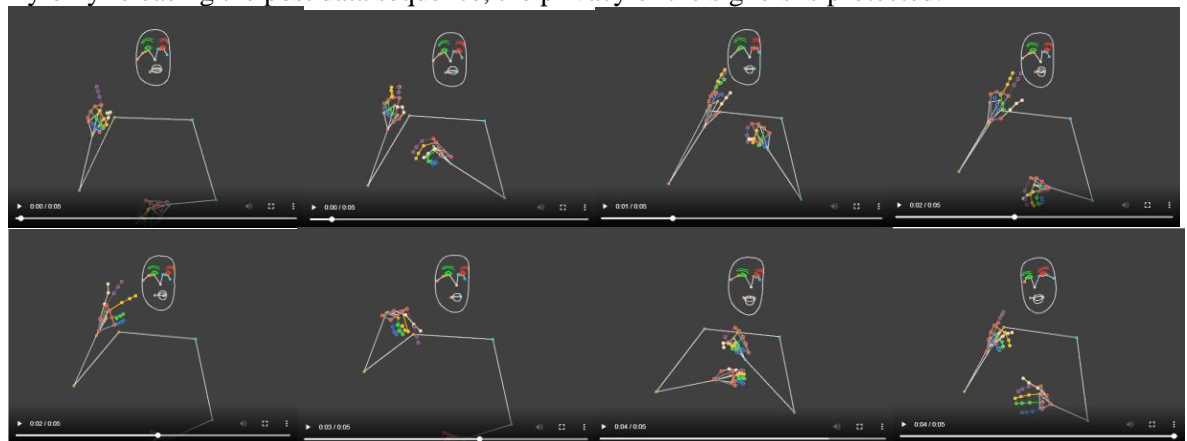

Figure 1. ASL signing sequence for sentence "the change will happen in the third quarter of this year"

### 3.2.3. Data Correction

No matter how robust its algorithms, a fully automatic system will always have errors and mistakes. It is important to have a user-friendly system to allow human intervention. A web-based interface, as shown in Figure 2, was developed to allow users to review the dataset and to make minor corrections, for errors like OCR mistakes. As the closed caption regions in the ASL News contents all have an opaque border, such OCR mistakes are rare. The interface supports some simple features like marking bad sentences to be excluded from the dataset generation, and text editing to change the English annotation.

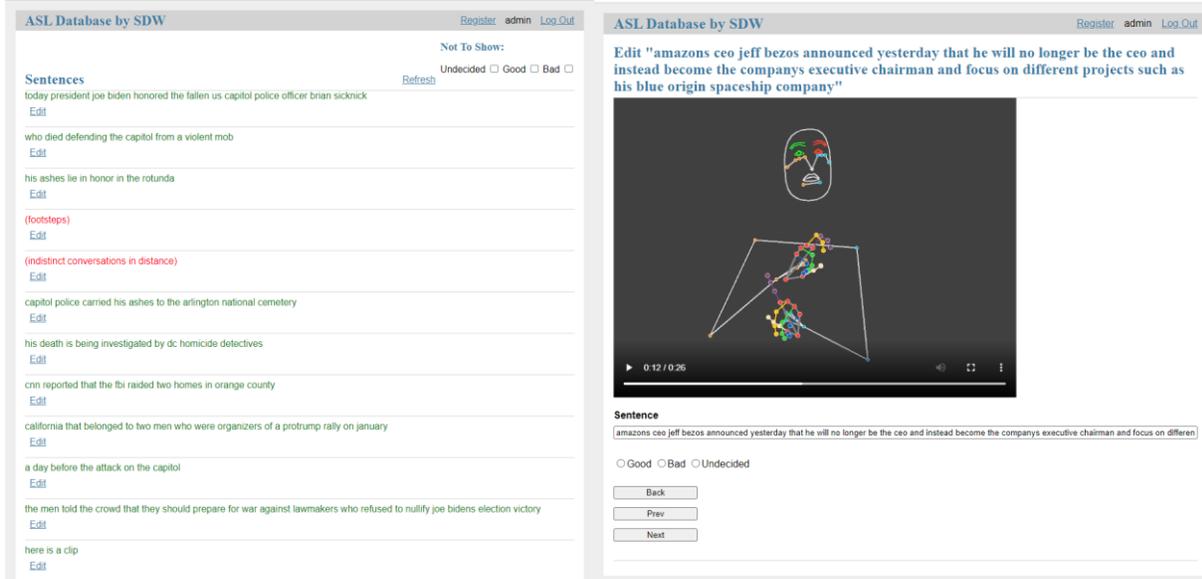

(a) (b)

Figure 2. An interactive web interface is provided to browse, review, and edit the SDW-ASL database. (a) shows the sentence browse view with the following color coding: red - bad, green - good, blue - undecided; (b) shows the viewing/editing window for a selected sentence.

### 3.3. SDW-ASL Database System and Dataset

The raw ASL video data has been collected from two ASL News channels. One channel has one male anchor. The other channel is primarily covered by one male anchor with two female anchors substituting on a regular basis. Only the regular news clips with consistent settings are used. Other content types such as interviews, event reports, holiday celebrations are not used. Furthermore, only the ones with either WebVTT or video-embedded closed captions are used. Over 690 video clips have been collected, with over 100 hours of raw video, yielding over 92 hours of annotated ASL data. These clips are an average of 9.5 minutes long.

Based on the data collected so far, the first SDW-ASL dataset generated contains a total of 92 hours of ASL video with over 416k English words. These words form a vocabulary of 18k tokens. Distribution of the vocab tokens is shown in Figure 3.0. The dataset contains various sizes of sentences. The example shown in Figure 1 has a 5-second short sentence with 11 English words. On the other hand, the example in Figure 2 is a 26-second-long sentence with 33 English words.

Data is stored and processed on a desktop computer with Intel 11th generation i5 Core with 6 cores running at 2.6GHz frequency.

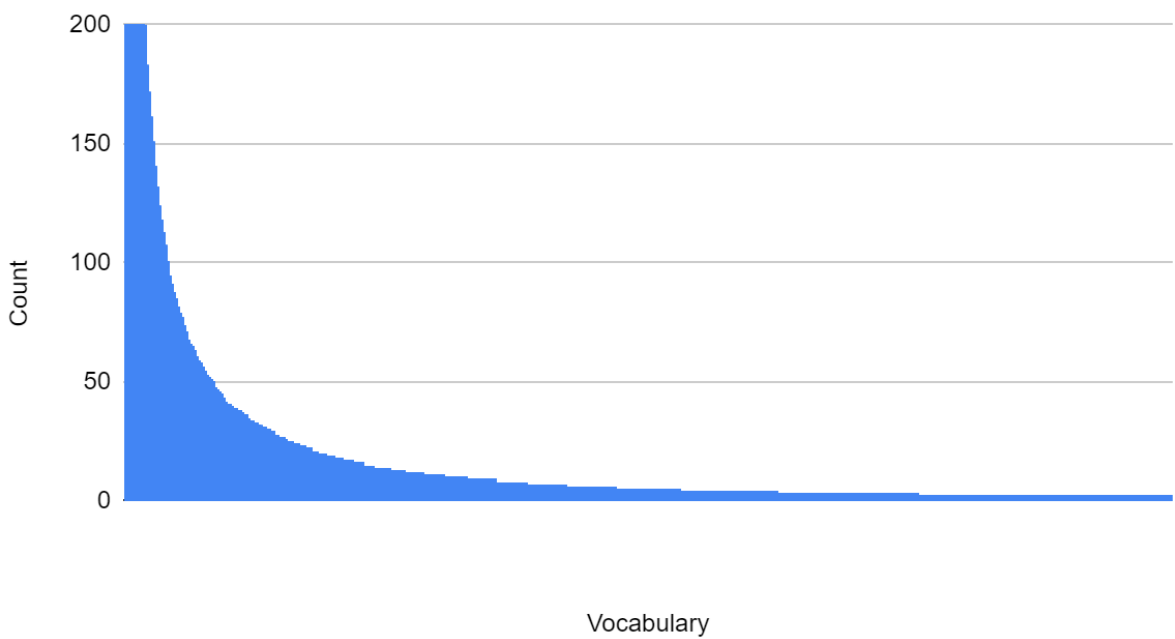

Figure 3. Vocab distribution in SDW-ASL dataset.

## 4. Conclusion and Future Work

We proposed a system that can generate large scale ASL datasets for continuous ASL. It is suitable for general ASL processing and is particularly useful for ASL production. The continuous ASL dataset contains English labeled human actuations in condensed body pose data formats. We are releasing the first version of the ASL dataset to better serve the research community. It contains 30k sentences, or 416k words with a vocabulary of 18k words in a total of 104 hours. This is the largest continuous sign language dataset published to date, in terms of video duration. We also describe a system that can evolve and expand the dataset to incorporate better data processing techniques and more contents when available. It is our hope that the release of this ASL dataset and the sustainable dataset generation system to the public will propel better deep-learning research in ASL natural language processing.